\documentclass[10pt,twocolumn,letterpaper]{article}
\usepackage{cvpr}
\usepackage{times}
\usepackage{epsfig}
\usepackage{graphicx}
\usepackage{amsmath}
\usepackage{amssymb}
\usepackage{bm}
\usepackage{enumerate}
\usepackage{color}
\usepackage{multirow}
\usepackage{array}
\usepackage{bigstrut}
\usepackage{algorithmic}
\usepackage{algorithm}
\usepackage{arydshln}

\usepackage[caption=false,font=footnotesize]{subfig}
\graphicspath{{figures/}}

\usepackage[pagebackref=true,breaklinks=true,letterpaper=true,colorlinks,bookmarks=false]{hyperref}

\newcommand{\comment}[1]{}
\newcommand{\deflen}[2]{%
    \expandafter\newlength\csname #1\endcsname
    \expandafter\setlength\csname #1\endcsname{#2}%
}

\cvprfinalcopy 

\def\cvprPaperID{1078} 

\ifcvprfinal\fi
\begin{document}

\title{Zoom and Learn: Generalizing Deep Stereo Matching to Novel Domains}

\author{Jiahao Pang$^1$\hspace{9pt}Wenxiu Sun$^1$\hspace{9pt}Chengxi\,Yang$^1$\hspace{9pt}Jimmy\,Ren$^1$\hspace{9pt}Ruichao\,Xiao$^1$\hspace{9pt}Jin\,Zeng$^1$\hspace{9pt}Liang\,Lin$^{1, 2}$\\
$^1$SenseTime Research\hspace{30pt}$^2$Sun Yat-sen University\\
{\tt\small \{pangjiahao,\hspace{3pt}sunwenxiu,\hspace{3pt}yangchengxi,\hspace{3pt}rensijie,\hspace{3pt}xiaoruichao,\hspace{3pt}zengjin,\hspace{3pt}linliang\}@sensetime.com}
}

\maketitle

\begin{abstract}
Despite the recent success of stereo matching with convolutional neural networks (CNNs), it remains arduous to generalize a pre-trained deep stereo model to a novel domain. 
A major difficulty is to collect accurate ground-truth disparities for stereo pairs in the target domain. 
In this work, we propose a self-adaptation approach for CNN training, utilizing both synthetic training data (with ground-truth disparities) and stereo pairs in the new domain (without ground-truths). 
Our method is driven by two empirical observations. 
By feeding real stereo pairs of different domains to stereo models pre-trained with synthetic data, we see that: i) a pre-trained model does not generalize well to the new domain, producing artifacts at boundaries and ill-posed regions; however, ii) feeding an up-sampled stereo pair leads to a disparity map with extra details. 
To avoid i) while exploiting ii), we formulate an iterative optimization problem with graph Laplacian regularization. 
At each iteration, the CNN adapts itself better to the new domain: we let the CNN learn its own higher-resolution output; at the meanwhile, a graph Laplacian regularization is imposed to discriminatively keep the desired edges while smoothing out the artifacts. 
We demonstrate the effectiveness of our method in two domains: daily scenes collected by smartphone cameras, and street views captured in a driving car.
\end{abstract}
\vspace{-10pt}

\vspace{-10pt}
\section{Introduction}
\label{sec:intro}
Stereo matching is a classic yet important problem for many computer vision tasks ({\it e.g.,} 3D reconstruction \cite{geiger2011stereoscan} and autonomous vehicles \cite{geiger2012we}). 
Particularly, given a rectified image pair captured by stereo cameras, one aims at estimating the disparity of each pixel between the two images. 
Traditionally, a stereo matching pipeline starts from matching cost computation and cost aggregation. 
Further optimization and refinement lead to the output disparity \cite{hirschmuller2008stereo}. 
Recent advances in deep learning has inspired a lot of end-to-end convolutional neural networks (CNNs) for stereo matching, {\it e.g.,} \cite{kendall2017end,mayer2016large}. 
Unlike the traditional wisdom, an end-to-end CNN integrates the stereo matching pipeline into a holistic deep architecture by learning from the training data.
Under confined scenarios with proper training data ({\it e.g.,} the KITTI dataset \cite{geiger2012we}), the end-to-end deep stereo models achieve unprecedented state-of-the-art performance.

However, it remains difficult to generalize a pre-trained deep stereo model to a novel scenario.
Firstly, the contents in the source domain may have very different characteristics from the target domain. 
Moreover, real stereo pairs collected with different stereo modules suffer from several degenerations---{\it e.g.}, noise corruption, photometric distortions, imperfections in rectification---to different extents.
Directly feeding a stereo pair of the target domain to a CNN pre-trained from another domain deteriorates its performance significantly. 
Consequently, state-of-the-art approaches, {\it e.g.}, \cite{kendall2017end,pang2017cascade}, train their models with synthetic datasets \cite{mayer2016large}, then perform finetuning on a fewer amount of domain-specific data with ground-truths. 
Unfortunately, besides a few public datasets for research purpose, {\it e.g.}, the KITTI dataset \cite{geiger2012we} and the Middlebury dataset \cite{scharstein2014high}, it is expensive and troublesome to collect real stereo pairs with accurate ground-truth disparities.

To resolve this dilemma, we propose a self-adaptation approach to generalize deep stereo matching methods to novel domains. 
We utilize synthetic training data and stereo pairs of the target domain, where only the synthetic data have known disparity maps.
Our approach is compatible with end-to-end deep stereo methods, {\it e.g.,} \cite{mayer2016large,pang2017cascade}, guiding a pre-trained model to gradually adapt to the target scenario. 
We start our explorations by feeding real stereo pairs from different domains to models pre-trained with synthetic data, resulting in two empirical observations: 
\begin{enumerate}[(i)]
\item {\emph{Generalization glitches}: a pre-trained model does not generalize well on the target domain---the produced disparity maps can be blurry at object edges and erroneous at ill-posed regions;}
\item{\emph{Scale diversity}: feeding a properly up-sampled stereo pair (the same stereo pair at a finer scale) leads to another disparity map with more meaningful details, {\it e.g.}, sharper object boundaries, more high-frequency contents of the scene.}
\end{enumerate}
To avoid the issues of (i) while exploiting the benefits of (ii), we propose an iterative regularization scheme for finetuning deep stereo matching models. 

\begin{figure}[!t]
        \centering
        \subfloat{\includegraphics[width=\figobsA]{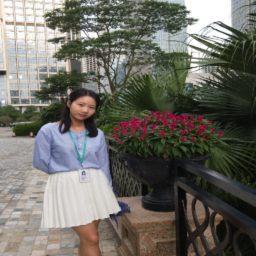}}\hspace{0pt}
        \subfloat{\includegraphics[width=\figobsA]{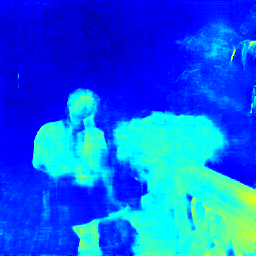}}\hspace{0pt}
        \subfloat{\includegraphics[width=\figobsA]{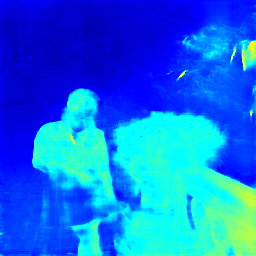}}\hspace{0pt}
        \subfloat{\includegraphics[width=\figobsA]{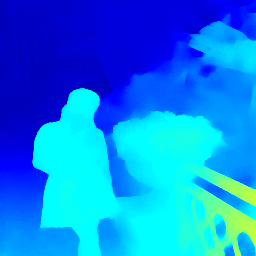}}\hspace{0pt}\\ \vspace{-5pt}
        \subfloat{\includegraphics[width=\figobsA]{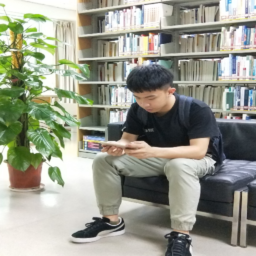}}\hspace{0pt}
        \subfloat{\includegraphics[width=\figobsA]{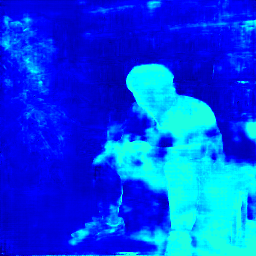}}\hspace{0pt}
        \subfloat{\includegraphics[width=\figobsA]{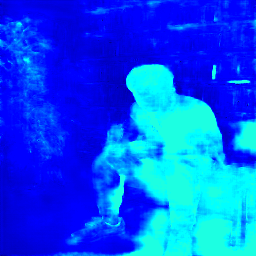}}\hspace{0pt}
        \subfloat{\includegraphics[width=\figobsA]{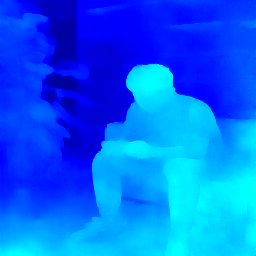}}\hspace{0pt}\\
        \addtocounter{subfigure}{-10}\vspace{-5pt}
        \subfloat[Left]{\includegraphics[width=\figobsA]{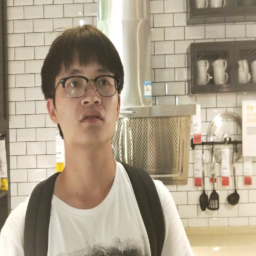}}\hspace{0pt}
        \subfloat[DispNetC]{\includegraphics[width=\figobsA]{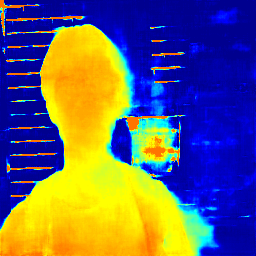}}\hspace{0pt}
        \subfloat[DispNetC-80]{\includegraphics[width=\figobsA]{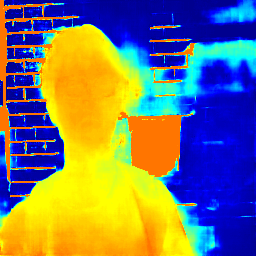}}\hspace{0pt}
        \subfloat[Ours]{\includegraphics[width=\figobsA]{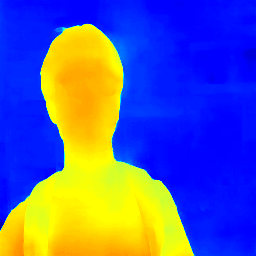}}\\
    \caption{Feeding stereo pairs collected from smartphones to models pre-trained from synthetic data leads to blurry edges and artifacts. In contrast, our self-adaptation approach brings significant improvements to the disparity maps.}
    \label{fig:obsA}
\end{figure}

We formulate the CNN training as an iterative optimization problem with graph Laplacian regularization. 
On one hand, we let the CNN learn its own finer-grain output; on the other hand, a graph Laplacian regularization is imposed to discriminatively retain the useful edges while smoothing out the undesired artifacts. 
Our formulation, composing of a data term and a smoothness term, is solved iteratively, leading to a model well suited for the novel domain {\it e.g.}, Figure\,\ref{fig:obsA}. 
The proposed self-adaptation approach is called {\it zoom and learn}, or ZOLE, for short. 
We demonstrate the effectiveness of our approach to two different domains: daily scenes collected by smartphone cameras, and street views captured from the perspective of a driving car.

This paper is structured as follows. 
Related works are reviewed in Section\,\ref{sec:related}. 
We then illustrate our observations about deep stereo models in Section\,\ref{sec:observe}. 
The proposed self-adaptation approach is introduced in Section\,\ref{sec:iter}. 
Section\,\ref{sec:results} presents the experimental results and Section\,\ref{sec:conclusion} concludes our work.

\section{Related Works}
\label{sec:related}
\deflen{figobsA}{56pt} 
We first review several stereo matching algorithms based on convolutional neural networks (CNNs).
We then turn to related works on graph Laplacian regularization and iterative regularization/filtering.

\textbf{Deep stereo algorithms}: 
Recent breakthroughs in deep learning have reshaped the paradigms of many computer vision tasks, including stereo matching. 
Early works employing CNNs for stereo matching focuses on learning a robust similarity measure for matching cost computation {\it e.g.}, \cite{han2015matchnet,zbontar2016stereo}. 
To produce disparity maps, modules in the traditional stereo matching pipeline are indispensable.
The remarkable work, \emph{DispNet}, proposed by Mayer {\it et al}~\cite{mayer2016large}, is the first end-to-end CNN approach for stereo matching, where an encoder-decoder architecture is employed for supervised learning. 
Other recent works with leading performance include CRL~\cite{pang2017cascade}, GC-NET~\cite{kendall2017end}, DRR~\cite{gidaris2016detect}, {\it etc}. 
These works explore different CNN architectures tailor-made for stereo matching. 
They achieve superior results on the KITTI 2015 stereo benchmark \cite{geiger2012we}, a benchmark containing driving scenes. 
Despite the success of these methodologies, to adopt them in a novel domain, it is necessary to fine-tune the models with new domain-specific data. 
Unfortunately, in practice, it is very difficult to collect accurate disparity maps for training \cite{geiger2012we,scharstein2014high}. 

To mitigate this problem, some recent works proposed semi-/un-supervised approaches to train a CNN model for stereo matching (or its related problem, monocular depth estimation).
This category of works is essentially based on left-right consistency/warping, {\it e.g.}, \cite{godard2017unsupervised,kuznietsov2017semi,zhong2017self,zhou2017unsupervised}. 
For instance, one may synthesize the left (or right) view according to the estimated left (or right) disparity and the right (or left) view for computing a loss function. 
However, left-right consistency becomes vulnerable when the stereo pairs are imperfect, {\it e.g.}, when the two views have different photometric distortions. 
Another line of research by Tonioni~{\it et~al.}~\cite{tonioni2017unsupervised} propose to finetune a pre-trained model to achieve domain adaptation. 
Their method relies on the results of other stereo methods and confidence measures. 
Our work also performs finetuning with a pre-trained stereo model. 
In contrast, we do not rely on external models or setups: our self-supervised domain adaptation method lets the CNN discriminatively learn the useful details from its own finer-grain outputs.

\textbf{Other related works}: 
According to \cite{liu2017random,shuman2013emerging}, graph Laplacian regularization is particularly useful for the recovery of piecewise smooth signals, {\it e.g.}, disparity maps. 
By having an appropriate graph, edges can be preserved while undesired defects are suppressed \cite{pang2017graph, pang2015optimal}. 
Hence, we propose to apply graph Laplacian regularization to selectively learn and preserve the meaningful details from the higher-resolution disparity outputs.

Iterative regularization/filtering is an important technique in classic image restoration \cite{katsaggelos1989iterative,milanfar2013tour,osher2005iterative}. 
To restore a corrupted image, it is regularized iteratively through a variational formulation, so that its quality improves at each iteration. 
To utilize scale diversity while avoiding generalization glitches (as mentioned in Section\,\ref{sec:intro}), we embed iterative regularization into the CNN training process, making the model parameters improve gradually. 
Different from iterative refinement via a stacked neural network architecture, {\it e.g.}, \cite{ilg2017flownet,ummenhofer2017demon}, our iterative process occurs during training.

\section{Observations}
\label{sec:observe}
We first present two phenomena by feeding real-world stereo pairs in different domains to deep stereo models pre-trained with synthetic datasets ({\it e.g.}, FlyingThings3D \cite{mayer2016large}, MPI Sintel \cite{butler2012naturalistic}, Virtual KITTI \cite{gaidon2016virtual}). 
Underlying reasons for these phenomena will also be presented.
We choose the off-the-shelf \emph{DispNet} \cite{mayer2016large} architectures---both the one with explicit correlation (DispNetC) and the one based on convolution only (DispNetS)---for our discussions. 
Their encoder-decoder architectures are representative and also widely used in the deep learning literature, {\it e.g.}, \cite{badrinarayanan2017segnet,long2015fully,ronneberger2015u}.
\subsection{Generalization Glitches}\label{ssec:glitch}
In general, a stereo model pre-trained with synthetic data does not perform well on real stereo data in a particular domain. 
Firstly, the contents of the synthetic data may differ from that of the target domain. 
Moreover, real stereo pairs inevitably suffer from defects arising from the imaging process. 
For instance, they are likely corrupted by noise. 
Besides, the two views may have different photometric distortions due to inconformity of the two cameras.
In some cases, the stereo pair may not even be well rectified, {\it e.g.}, two corresponding pixels are not on the same scan-line. 
All the above factors deteriorate the performance of a model pre-trained with synthetic data.

For illustration, we use smartphones equipped with two rear-facing cameras to collect a few stereo pairs (of size 1024$\times$1024), then perform the following tests. 
We first adopt the released DispNetC model pre-trained with the FlyingThings3D dataset \cite{mayer2016large}. 
Since stereo pairs of smartphones have small disparity values, we also finetune a model from the released model, where we remove those FlyingThings3D stereo pairs with maximum disparity larger than 80. 
Data augmentation is introduced for the two views individually during training, please refer to Section\,\ref{sec:results} for more details. 
The resulting model is called DispNetC-80. 
Both DispNetC and DispNetC-80 perform very well on the FlyingThings3D dataset, but are problematic when applied to real smartphone data. 
Figure\,\ref{fig:obsA} shows a few disparity estimates of DispNetC and DispNetC-80. 
As can be seen, the results are blurry at object edges. 
Moreover, at ill-posed regions, {\it i.e.}, object occlusions, repeated patterns, and textureless regions, the disparity maps are erroneous.
In this work we call this \emph{generalization glitches}, meaning the mistakes that a deep stereo model (pre-trained with synthetic data) make when it is applied to real stereo pairs of a certain domain.
\begin{figure}
        \centering
        \subfloat{\includegraphics[width=20.5pt]{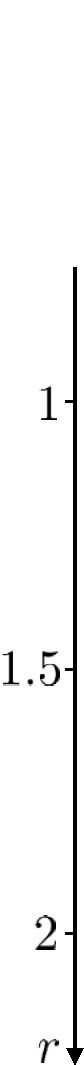}}\hspace{0pt}
        \subfloat{\includegraphics[width=210pt]{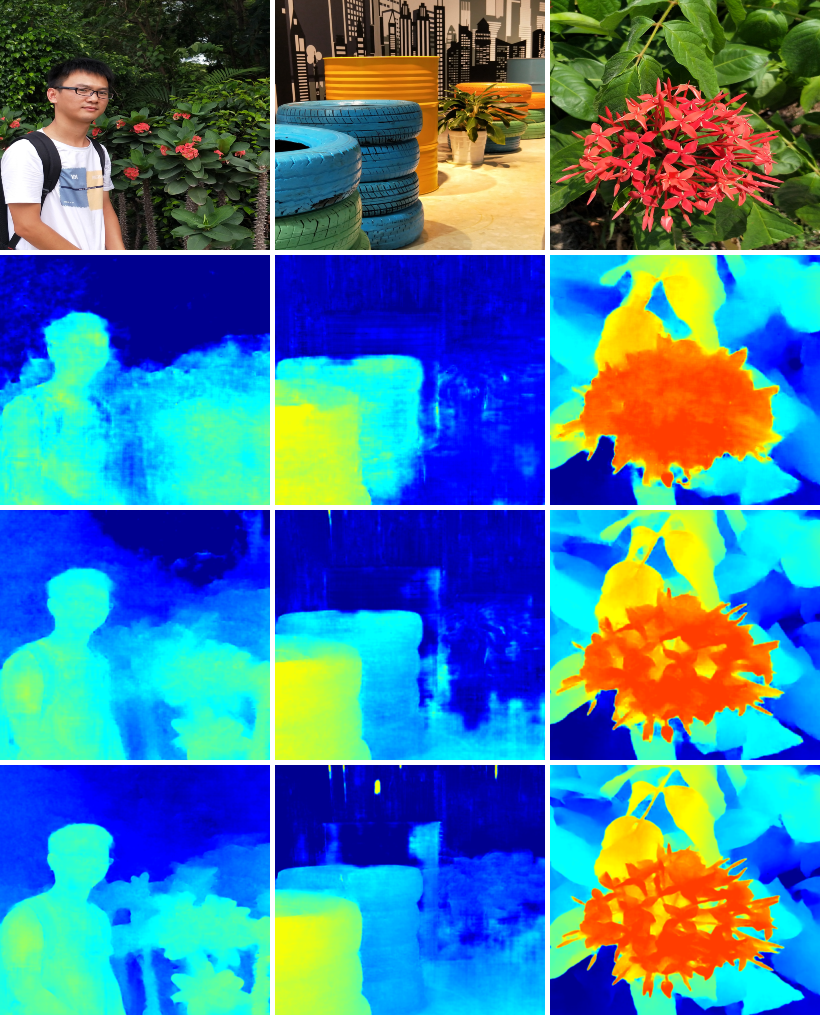}}
    \caption{For the same stereo pair, feeding its zoomed-in version to a stereo matching CNN leads to a disparity map with extra details. The four rows are the left image, the disparity maps obtained by \eqref{eq:finer}, with up-sampling ratio $r=1, 1.5, 2$, respectively.}
    \label{fig:obsB}
\end{figure}
\subsection{Scale Diversity}\label{ssec:scale}
In spite of the unpleasant generalization glitches, we find that deep stereo models have an encouraging property. 
Suppose we have a stereo pair $P = (L,R)$, where $L$ and $R$ are the left and the right views, respectively. 
We denote a deep stereo model parameterized by ${\bf{\Theta }}$ as $S(\bm\cdot;{\bf{\Theta }})$. 
By applying it to the stereo pair $P$ leads to a disparity map $D = S\left( {P;{\bf{\Theta }}} \right)$. 
The operation of up-sampling by $r$ times is denoted as ${\uparrow_r}(\bm\cdot)$， while down-sampling by $r$ times is ${\downarrow_r}(\bm\cdot)$. 
By passing an up-sampled stereo pair to $S$ then down-sampling the result, we obtain another disparity map, $D'$, of the same size as $D$,
\begin{equation}\label{eq:finer}
    D' = \frac{1}{r}\cdot{ \downarrow _r}\left( {S\left( {{ \uparrow _r}(P);{\bf{\Theta }}} \right)} \right).
\end{equation}
Note that after downsampling, the factor $1/r$ is necessary for making $D'$ to have the correct scaling. 
Compared to $D$, $D'$ usually contains more high-frequency details. 
To see this, we apply the released DispNetC model to a few stereo pairs captured by smartphones. 
We make the original size of the stereo pairs as $640\times 640$. 
For each of them, we estimate three disparity maps based on \eqref{eq:finer} with $r\in\{1, 1.5, 2\}$. Visual results are shown in Figure\,\ref{fig:obsB}. 
We see that as $r$ grows, more fine details are produced on the disparity maps. 

\begin{table}[t!]
  \small
  \centering
  \caption{The average three-pixel error rates of the released DispNetC and DispNetS models on the training set of KITTI stereo 2015. A resolution of $N$ means the stereo pairs are resized to $N\times N$ before passing to the CNNs.}\vspace{2pt}
    \begin{tabular}{l||c|c|c|c|c}
    \hline
    \multicolumn{1}{c||}{\multirow{2}[4]{*}{Network}} & \multicolumn{5}{c}{\vspace{-1.5pt}$\phantom{\hat{I}}\mathop{\textrm{ Resolution}}\limits_{\phantom{.}}\phantom{\hat{I}}$} \bigstrut\\
\cline{2-6}          & 896   & 1280   & 1664  & 2048  & 2432 \bigstrut[t]\\
    \hline
    DispNetC & 14.26\% & 9.97\% & {\bf 8.81}\% & 9.17\% & 10.53\% \bigstrut[t]\\
    DispNetS & 18.95\% & 11.61\% & 9.18\% & {\bf 8.64}\% & 9.08\% \bigstrut[t]\\
    \hline
    \end{tabular}%
  \label{tab:obsB}%
\end{table}%

However, a bigger $r$ does not necessarily mean better results. 
For further inspection, we adopt the released DispNetC and DispNetS models (trained with the FlyingThings3D dataset) and measure their performance on the training set of KITTI stereo 2015 \cite{geiger2012we} at different resolutions. 
The results, in terms of the percentage of pixels with an error greater than 3, or three-pixel error rate (3ER), are listed in Table\,\ref{tab:obsB}. 
We see that as the input resolution increases, the performance first improves then deteriorates. Because: 
\begin{enumerate}[(i)]
    \item{Up-sampling the stereo pairs enables the model to perform stereo matching at a localized manner with subpixel accuracy. Hence, more details on the stereo pairs are taken into account for computation, leading to disparity estimates with extra high-frequency contents;}
    \item{A finer-scale input translates to a smaller effective search range (or receptive field). As a CNN becomes too ``short-sighted,'' it lacks non-local information to estimate a proper disparity map, and its performance start to decline.}
\end{enumerate}
This phenomenon---different results can be observed with different input scales---is called \emph{scale diversity}, akin to the concept of transmit diversity in communication \cite{rappaport1996wireless}. 
We find that scale diversity also exists in other problems, {\it e.g.}, optical flow estimation \cite{ilg2017flownet,mayer2016large} and image segmentation \cite{long2015fully}, please refer to the supplementary material for more details.

\section{Zoom and Learn}
\label{sec:iter}
To achieve effective self-adaptation, our approach---zoom and learn (ZOLE)---finetunes a model pre-trained with synthetic data. 
It iteratively suppresses generalization glitches while utilizing the benefits of scale diversity.

\subsection{Graph Laplacian Regularization}
Graph Laplacian regularization is employed in a wide range of image restoration literature, {\it e.g.}, \cite{elmoataz2008nonlocal,gilboa2007nonlocal,milanfar2013tour}. 
It is also proven to be effective for the recovery of piecewise smooth signals \cite{hu2016graph,pang2017graph,shuman2013emerging}. 
We adopt graph Laplacian regularization (on a patch-by-patch basis) to guide the learning of CNNs. 
Graph Laplacian regularization assumes the ground-truth signal ${\bf s}\in \mathbb{R}^m$---in our case, a patch on the ground-truth disparity---is smooth with respect to a predefined graph ${\cal G}$ with $m$ vertices. 
Specifically, it imposes that the value of ${\bf s}^{\rm T}{\bf L}{\bf s}$, {\it i.e.}, the graph Laplacian regularizer, should be small for the ground-truth patch ${\bf s}$, where ${\bf L}\in \mathbb{R}^{m\times m}$ is the \emph{graph Laplacian matrix} of graph ${\cal G}$.
Given a disparity map $D$ produced by a deep stereo model, we compute the values of the graph Laplacian regularizers for the patches on $D$. 
The obtained values are summed up as a graph Laplacian regularization loss for CNN training. 

For an effective regularization with graph Laplacian, it is critical to constructing a graph ${\cal G}$ properly. 
We employ the graph structure of \cite{hein2007graph,pang2017graph} which works well for disparity map denoising. 
For illustration, we first introduce the concept of \emph{exemplar patches}. 
Exemplar patches are a set of $K$ patches, ${\bf f}_k\in\mathbb{R}^{m}$ where $1\le k\le K$, that are statistically related to the ground-truth patch ${\bf s}$. 
For instance, an exemplar patch can be a rough estimate of ${\bf s}$, or the co-located patch on the left image, {\it etc}. 
Our choices of the exemplar patches will be presented in Section\,\ref{ssec:iter}. 
With the exemplar patches, the edge weight $w_{ij}$ connecting pixel $i$ and pixel $j$ on patch ${\bf s}$ is given by
\begin{equation*}
w_{ij} = \left\{ {\begin{array}{*{20}{l}}
{\exp \left( { - \displaystyle{{d_{ij}^2}}} \right)} & {\mbox{if} ~~ \left| d_{ij} \right| \le \epsilon,}\\
0&{{\rm{otherwise,}}}
\end{array}} \right.
\end{equation*}
where $\epsilon$ is a threshold, $d_{ij}^2$ is a distance measure between pixel $i$ and pixel $j$. 
Hence, the resulting graph $\cal G$ is an $\epsilon$-neighborhood graph, {\it i.e.}, there is no edge connecting two pixels with a distance greater than $\epsilon$. 
We choose an individual value of $\epsilon$ for each patch, making every vertex of the graph has at least 4 edges.
The distance measure $d_{ij}^2$ is defined as follows:
\begin{equation}\label{eq:graph_rest}
d_{ij}^2= \sum\nolimits_{k = 1}^K{\left ( {\bf f}_k(i) - {\bf f}_k(j) \right )^2} + \alpha\bm\cdot l_{ij}^2,
\end{equation}
where ${\bf f}_k(i)$ and ${\bf f}_k(j)$ denote the $i$-th and the $j$-th entries of ${\bf f}_k$, respectively, so the first term of \eqref{eq:graph_rest} measures the Euclidean distance between pixels $i$ and $j$ in a $K$-dimensional space defined by the exemplar patches. 
$l_{ij}$ is simply the spatial distance (length) between pixels $i$ and $j$, and $\alpha$ is a constant weight, empirically set to be a small value $0.2$.

The adjacency matrix of ${\cal G}$ is denoted as ${\bf A}$, where the $(i,j)$-th entry of ${\bf A}$ is $w_{ij}$. 
The degree matrix of ${\cal G}$ is a diagonal matrix ${\bf D}$, its $i$-th diagonal entry is $\sum\nolimits_{j=1}^{m}w_{ij}$. 
Then the graph Laplacian ${\bf L}$ is given by ${\bf L} = {\bf D} - {\bf A}$, leading to the graph Laplacian regularizer ${\bf s}^{\rm T}{\bf L}{\bf s}\in\mathbb{R}$. 
From the analysis of \cite{pang2017graph}, graph Laplacian regularizer is an \emph{adaptive metric}. 
If the same edge (or gradient ) pattern appears in the majority of the exemplar patches, minimizing the graph Laplacian regularizer promotes the very edge pattern; if the exemplar patches are inconsistent, graph Laplacian regularization leads to a smoothed patch. 
We exploit this property to guide a deep stereo model to selectively learn the desired details.

\subsection{Training by Iterative Regularization}\label{ssec:iter}
We borrow the notion of iterative regularization \cite{milanfar2013tour} for generalizing deep stereo models to novel domains, giving rise to the proposed zoom and learn approach. 
Suppose we have a deep stereo model $S(\bm\cdot;{\bf\Theta}^{(0)})$ (parameterized by ${\bf\Theta}^{(0)}$) pre-trained with synthetic data. 
We also have a set of $N$ stereo pairs, $P_i=(L_i, R_i), 1\le i\le N$, where the first $N_{\rm dom}$ of them are real stereo pairs of the target domain while the rest $N_{\rm syn} = N - N_{\rm dom}$ pairs are synthetic data, among which only the synthetic data has ground truth disparities $D_i$ ($N_{\rm dom} + 1\le i\le N$). 

We solve for a new set of model parameters ${\bf\Theta}^{(k+1)}$ at iteration $k$. 
For a constant $r>1$, we first create a set of ``ground-truths'' for the $N_{\rm dom}$ real stereo pairs by zooming (up-sampling), {\it i.e.},
\begin{equation}\label{eq:finer_iter}
{D_{i}} = \frac{1}{r}\bm\cdot{ \downarrow _r}\left( {S\left( {{ \uparrow _r}({P_i});{{\bf{\Theta }}^{(k)}}} \right)} \right), 1\le i\le N_{\rm dom}.
\end{equation}
From Section\,\ref{ssec:scale}, $D_{i}$ contains more details than $S(P_i; {\bf \Theta}^{(k)})$. 
We divide a disparity map $D_i$ into $M$ square patches tiling it where each patch is a vector of length $m$. The vectorization operator is denoted as ${\rm vec}(\bm\cdot)$ so that ${\rm vec}(D_i)\in\mathbb{R}^{Mm}$. The $m$-by-$Mm$ matrix extracting the $j$-th patch from $D_i$ is denoted as ${\bf R}_j$. With these settings, we formulate the following iterative optimization problem,
%
%
%
\begin{align}
{{\bf{\Theta }}^{(k + 1)}} &= \mathop {\arg \min }\limits_{\bf{\Theta }} \sum\limits_{i = 1}^{N_{\rm dom}}\hspace{-2pt}{\sum\limits_{j = 1}^M {{{\left\| {{{\bf{s}}_{ij}} - {\bf d}_{ij}} \right\|}_1}\hspace{-1pt}+\hspace{-1pt}\lambda \bm\cdot{\bf{s}}_{ij}^{\rm{T}}{\bf{L}}_{ij}^{(k)}{{\bf{s}}_{ij}}} +}\nonumber\\
&\hspace{47pt}\tau\bm\cdot\sum_{i={N_{{\rm{dom}}}} + 1}^N {{{\left\| {S({P_i};{\bf{\Theta }}) - {D_i}} \right\|}_1} },\label{eq:problem}\\
{\rm s.t.}\hspace{5pt} {{\bf{s}}_{ij}} &= {{\bf{R}}_j}\bm\cdot{\rm{vec}}\left( {S(P_i;{\bf{\Theta }})} \right), \hspace{5pt}{\bf d}_{ij} = {{\bf{R}}_j}\bm\cdot{\rm{vec}}\left( {D_i} \right).\nonumber
\end{align}
Here ${\bf s}_{ij}$ and ${\bf d}_{ij}$ are the $j$-th patches of ${S(P_i;{\bf{\Theta }})}$ and $D_i$, respectively. $\lambda$ and $\tau$ are positive constants. 
Our optimization problem $\eqref{eq:problem}$ first minimizes over each patch on the $N_{\rm dom}$ stereo pairs: the first term (data term) drives ${\bf s}_{ij}$ to be similar to ${\bf d}_{ij}$; and the second term (smoothness term) is a graph Laplacian regularizer induces from the matrix ${\bf L}_{ij}^{(k)}$. 
The third term of $\eqref{eq:problem}$ lets ${\bf\Theta}^{(k+1)}$ be a feasible deep stereo model; it literally means that: a deep stereo model works well for the target domain should also has reasonable performance on the synthetic data.

At iteration $k$, a graph ${\cal G}_{ij}^{(k)}$ ($1\le i\le N_{\rm dom}$, $1\le j\le M$), and hence the corresponding graph Laplacian, ${\bf L}_{ij}^{(k)}$, are pre-computed for calculating a loss ${\bf s}_{ij}^{\rm T}{\bf L}^{(k)}_{ij}{\bf s}_{ij}$. 
We choose the following three exemplar patches for building ${\cal G}_{ij}^{(k)}$:
\begin{align*}
    &{\bf f}_{\rm left} = w_{\rm left}\bm\cdot{\bf R}_j\bm\cdot{\rm vec}(L_i),\\
    &{\bf f}_{\rm curr} = w_{\rm curr}\bm\cdot{\bf R}_j\bm\cdot{\rm vec}(S(P_i;{\bf\Theta}^{(k)})),\\
    &{\bf f}_{\rm fine} = w_{\rm fine}\bm\cdot{\bf R}_j\bm\cdot{\rm vec}(D_i)=w_{\rm fine}\bm\cdot {\bf d}_{ij},
\end{align*}
where $w_{\rm left}$, $w_{\rm curr}$ and $w_{\rm fine}$ are constants. 
In other words, ${\bf f}_{\rm left}$, ${\bf f}_{\rm curr}$, and ${\bf f}_{\rm fine}$ are the $j$-th patches of the left image $L_i$, the current prediction $S(P_i;{\bf\Theta}^{(k)})$ and the finer-grain prediction $D_i$ \eqref{eq:finer_iter}, respectively.

Our chosen exemplar patches lead to a graph Laplacian regularizer that discriminatively retain the desired details from ${\bf f}_{\rm fine}$ whilst smoothing out possible artifacts on both ${\bf f}_{\rm curr}$ and ${\bf f}_{\rm fine}$. 
We analyze how the patches ${\bf f}_{\rm left}$, ${\bf f}_{\rm curr}$ and ${\bf f}_{\rm fine}$ affects the behavior of the graph Laplacian:
\begin{enumerate}[(i)]
    \item Suppose a desired object boundary (denoted by $A$) does not appear in the current predicted patch ${\bf f}_{\rm curr}$. However, it has appeared in the finer-grain patch ${\bf f}_{\rm fine}$ by virtue of scale diversity (Section\,\ref{ssec:scale}), then $A$ should also appear in ${\bf f}_{\rm left}$; otherwise the CNN cannot generate $A$ on ${\bf f}_{\rm fine}$. In this case, both ${\bf f}_{\rm fine}$ and ${\bf f}_{\rm left}$ have boundary $A$, resulting in a Laplacian ${\bf L}_{ij}^{(k)}$ that promotes $A$ on ${\bf s}_{ij}$.
    \item Suppose due to generalization glitches, an undesired pattern (denoted as $B$) is produced in one exemplar patch, ${\bf f}_{\rm curr}$ or ${\bf f}_{\rm fine}$. Since $B$ is absence in the other exemplar patches, the corresponding graph Laplacian ${\bf L}_{ij}^{(k)}$ penalizes $B$ on ${\bf s}_{ij}$.
\end{enumerate}
Hence, our graph Laplacian regularizer guides the CNN to only learn the meaningful details. 

\subsection{Practical Algorithm}\label{ssec:alg}
Iteratively solving the optimization problem \eqref{eq:problem} can be achieved by training the model $S(\bm\cdot;\bf\Theta)$ with standard backpropagation \cite{lecun2015deep}. 
We hereby present how to use the proposed formulation for finetuning a pre-trained model in practice. 
Since a disparity map $D_i$ is tiled by $M$ patches, ${\bf d}_{ij}$ with $1\le j\le M$, the first term in \eqref{eq:problem} equals $\sum\nolimits_{i=1}^{N_{\rm dom}} {{{\left\| {S({P_i};{\bf{\Theta }}) - {D_i}} \right\|}_1} }$. Hence, the objective of \eqref{eq:problem} can be rewritten as:
\begin{equation}\label{eq:problem_new}
\begin{split}
    &{{\bf{\Theta }}^{(k + 1)}} = \mathop {\arg \min }\limits_{\bf{\Theta }} \sum_{i=1}^{N_{\rm dom}} {{{\left\| {S({P_i};{\bf{\Theta }}) - {D_i}} \right\|}_1} } + \\
    &\tau\bm\cdot\sum_{i=N_{\rm dom}+1}^{N} {{{\left\| {S({P_i};{\bf{\Theta }}) - {D_i}} \right\|}_1} } + \lambda\bm\cdot\sum\limits_{i = 1}^{N_{\rm dom}} {\sum\limits_{j = 1}^M { {\bf{s}}_{ij}^{\rm{T}}{\bf{L}}_{ij}^{(k)}{{\bf{s}}_{ij}}} },
\end{split}    
\end{equation}
We see that the first two terms of \eqref{eq:problem_new} are simply L1 loss with different weightings for the target domain and the synthetic data. 
The third term is the proposed graph Laplacian regularization loss, we discuss its backpropagation in the supplementary material.

In general, there are a lot of training examples ($N$ is large), yet in practice, every training iteration can only take in a batch of $n\ll N$ training examples and perform stochastic gradient descent. 
As a result, we shuffle all the $N$ stereo pairs and sequentially take out $n$ of them to form a training batch for the current iteration. 
For a synthetic stereo pair $P_i$ ($N_{\rm dom}+1\le i\le N$) in the batch, we directly use its L1 loss for backpropagation since its ground-truth $D_i$ is known. 
Otherwise, for a stereo pair $P_i$ with $1\le i\le N_{\rm dom}$ in the batch, we first feed its up-sampled version to the CNN for computing the finer-grain ``ground-truth'' $D_i$, we also compute the current estimate $S(P_i; {\bf\Theta}^{(k)})$ and hence the graph Laplacian matrices ${\bf L}_{ij}^{(k)}$ for each patch. 
With $D_i$ and the pre-computed ${\bf L}_{ij}^{(k)}$'s, $1\le j\le M$, both the L1 loss and the graph Laplacian regularization loss are employed for backpropagation. 

For every $t$ training iterations, we perform a validation procedure with left-right consistency, using another set of $N_{\rm dom}^{\rm (v)}$ stereo pairs in the target domain. 
We first estimate the disparity maps with the up-to-date model then synthesize $N_{\rm dom}^{\rm (v)}$ left images with the estimated disparity maps and the right images. 
Then we compute the peak signal-to-noise ratios (PSNRs) between the synthesized left images and the genuine ones. 
The average PSNR reflects the performance of the current model. 
During the training process, we keep track of the best PSNR value ${v}^{(\rm bst)}$ and its corresponding model ${\bf\Theta}^{(\rm bst)}$. 
After $k_{\rm max}$ training iterations, we terminate the training and output ${\bf\Theta}^{(\rm bst)}$. 
Algorithm\,\ref{alg:iter} summarizes the key steps of our self-adaptation approach.
\begin{algorithm}[t]
\caption{Zoom and learn (ZOLE)}\label{alg:iter}
\begin{small}
\begin{algorithmic}[1]
\STATE {\bf{Input:}} Pre-trained deep stereo model $S(\bm\cdot;{\bf\Theta}^{(0)})$,\\
training data $\{P_i\}_{i=1}^{N_{\rm dom}}$ and $\{P_i, D_i\}_{i={N_{\rm dom}+1}}^{N}$
\STATE Shuffle the training data to form a list $\ell$
\FOR {$k = 0$ to $k_{\rm max} - 1$}
\FOR {$b=1$ to $n$}
\STATE Draw an index $i$ from list $\ell$
\IF {$i \le N_{\rm dom}$}
\STATE Compute $D_i$, then compute $S(P_i;{\bf\Theta}^{(k)})$ and hence the graph Laplacian matrices ${\bf L}_{ij}^{(k)}$ 
\ENDIF
\STATE Insert $\{P_i, D_i\}$ to the current batch
\ENDFOR
\STATE Use the formed training batch and the pre-computed Laplacian matrices to perform a step of gradient descent
\IF {${\rm mod}(k + 1, t)=0$}
\STATE Perform validation, update ${\bf\Theta}^{(\rm bst)}$ and ${v}^{(\rm bst)}$ if needed
\ENDIF
\ENDFOR
\STATE {\bf{Output:}} model parameters ${\bf\Theta}^{(\rm bst)}$
\end{algorithmic}
\end{small}
\end{algorithm}

\section{Experimentation}
\label{sec:results}
In this section, we generalize deep stereo matching for two different domains in the real world: daily scenes captured by smartphone cameras, and street views from the perspective of a driving car (the KITTI dataset \cite{geiger2012we}). 
We again choose the representative DispNetC \cite{mayer2016large} architecture for our experiments.
\subsection{Daily Scenes from Smartphones}\label{ssec:exp_phone}
Recently, many companies ({\it e.g.}, Apple, Samsung) have equipped their smartphones with two rear-facing cameras. 
Stereo pairs collected by these cameras have small disparity and possibly contaminated by noise due to the small area of their image sensors.
With two views of the same scene, stereo matching is applied to estimate a dense disparity map for subsequent applications, {\it e.g.}, synthetic bokeh \cite{Barron2015fast} and segmentation \cite{long2015fully}.
\begin{figure}
    \centering
    \includegraphics[width=220pt]{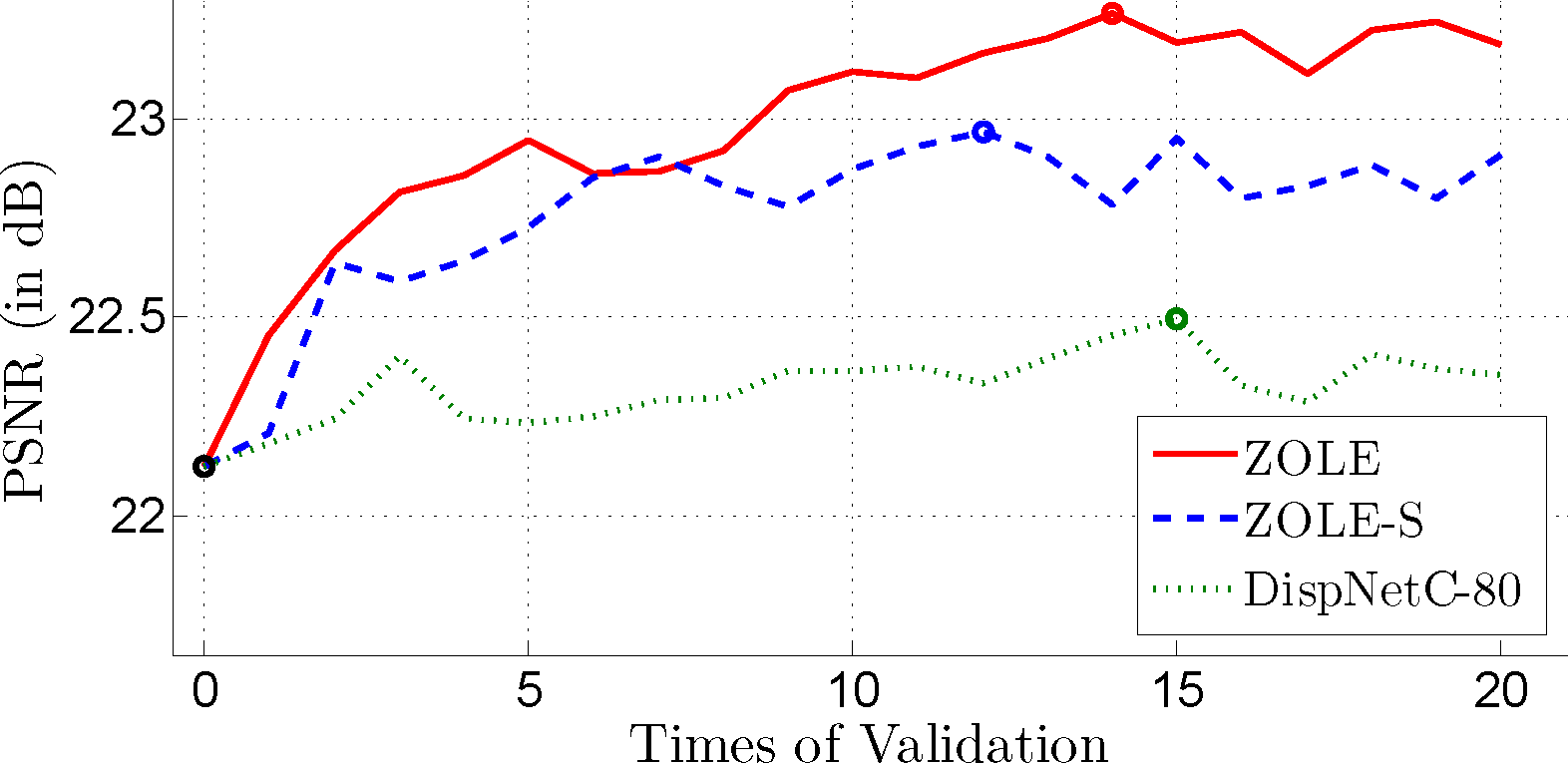}
    \caption{Validation performance of three different models during finetuning. The curves are plotted in terms of the average PSNR between the synthesized left images and the genuine ones.}
    \label{fig:evo}
\end{figure}

We aim at generalizing the released DispNetC model (pre-trained with the FlyingThings3D dataset \cite{mayer2016large}) for daily scenes captured by smartphones cameras.
For this purpose, we used various models of smartphones to collect $N_{\rm dom} = 1900$, $N_{\rm dom}^{\rm (v)} = 320$ and $N_{\rm dom}^{\rm (t)} = 320$ stereo pairs for training, validation, and testing, respectively. 
These stereo pairs contain daily scenes like human portraits and objects taken in various indoor and outdoor environments ({\it e.g.}, library, office, playground, park). 
All the collected images are rectified and resized to $768\times 768$, their ground-truth disparity maps are unknown.
Besides, we use the FlyingThings3D dataset for synthetic training examples in our method, they are also resized to $768\times 768$. 
Since their original size is $960\times 960$, their disparity maps need to be rescaled by a factor of 0.8. 
To cater for the small disparity values of the smartphone data, we only keep those synthetic examples with maximum disparity no greater than 80 after rescaling, leading to 9619 available examples. 
Among them, $N_{\rm syn}=8000$ examples are used for training and the rest $N_{\rm syn}^{\rm (t)}=1619$ are withheld for testing. 
We call this set of data FlyingThings3D-80.
In our experiments, all stereo pairs have intensity ranges from 0 to 255.

\deflen{figphone}{95pt}
\begin{figure*}[!t]
        \centering
        \subfloat{\includegraphics[width=\figphone]{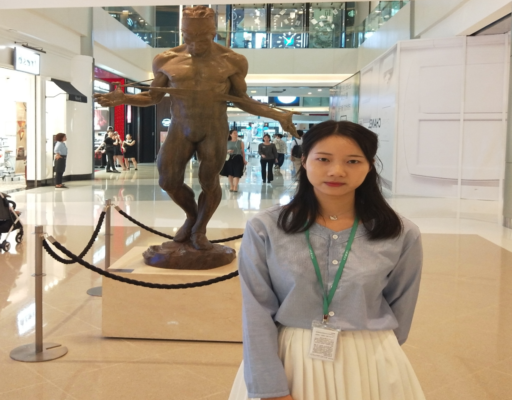}}\hspace{0pt}
        \subfloat{\includegraphics[width=\figphone]{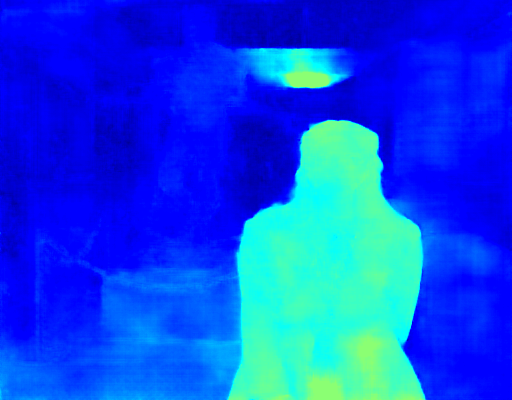}}\hspace{0pt}
        \subfloat{\includegraphics[width=\figphone]{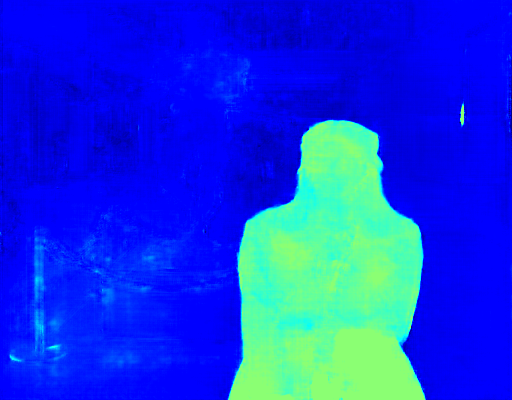}}\hspace{0pt}
        \subfloat{\includegraphics[width=\figphone]{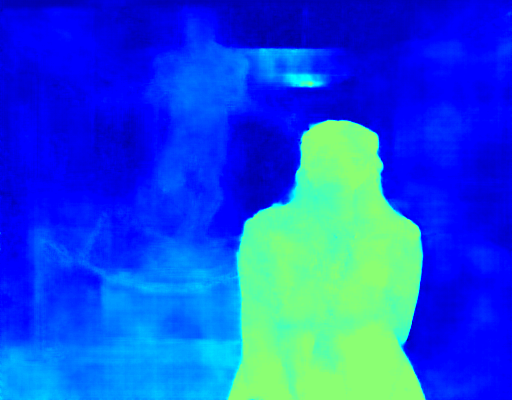}}\hspace{0pt}
        \subfloat{\includegraphics[width=\figphone]{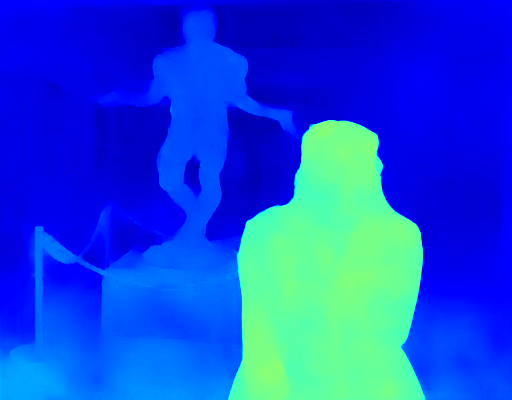}}\\ \vspace{-5pt}
        \subfloat{\includegraphics[width=\figphone]{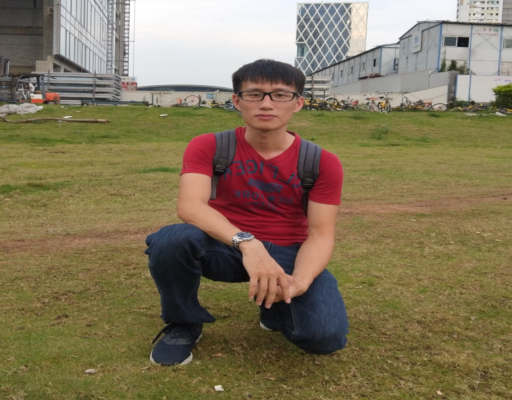}}\hspace{0pt}
        \subfloat{\includegraphics[width=\figphone]{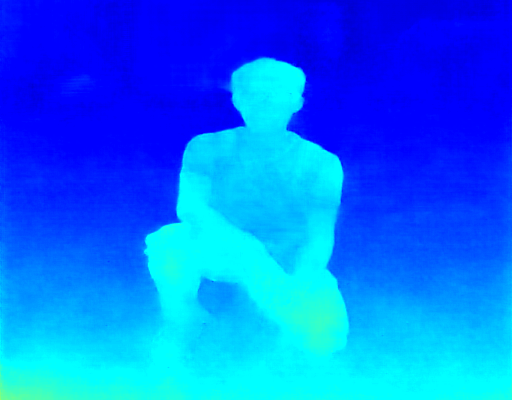}}\hspace{0pt}
        \subfloat{\includegraphics[width=\figphone]{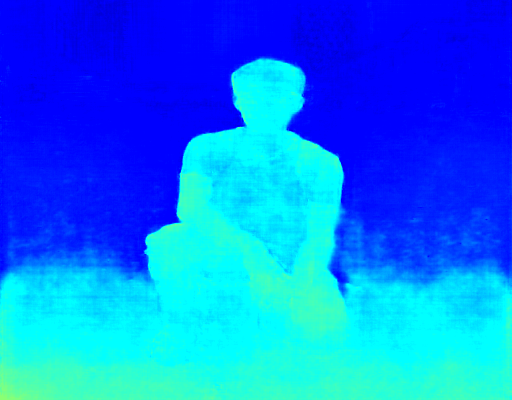}}\hspace{0pt}
        \subfloat{\includegraphics[width=\figphone]{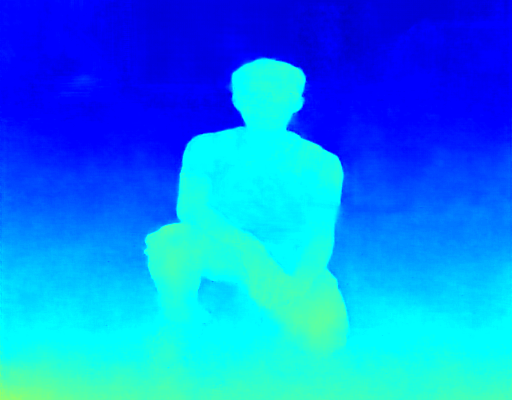}}\hspace{0pt}
        \subfloat{\includegraphics[width=\figphone]{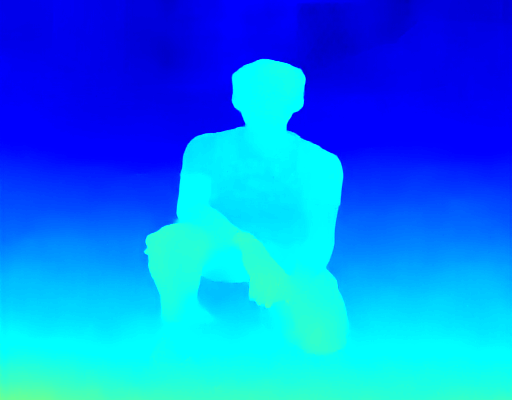}}\\
        \addtocounter{subfigure}{-10}\vspace{-5pt}
        \subfloat[Left image]{\includegraphics[width=\figphone]{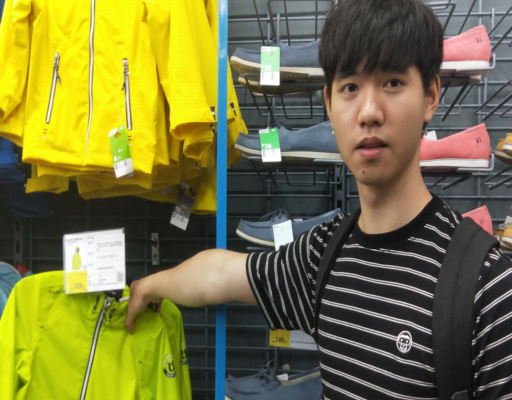}}\hspace{0pt}
        \subfloat[Tonioni {\it et al.}\,\cite{tonioni2017unsupervised}]{\includegraphics[width=\figphone]{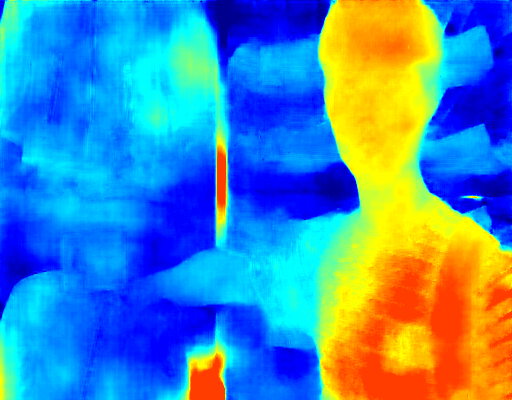}}\hspace{0pt}
        \subfloat[DispNetC]{\includegraphics[width=\figphone]{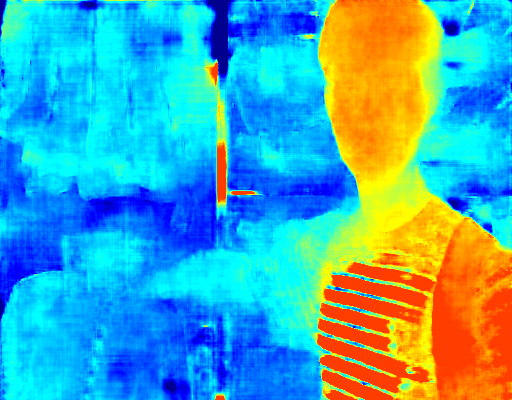}}\hspace{0pt}
        \subfloat[ZOLE-S]{\includegraphics[width=\figphone]{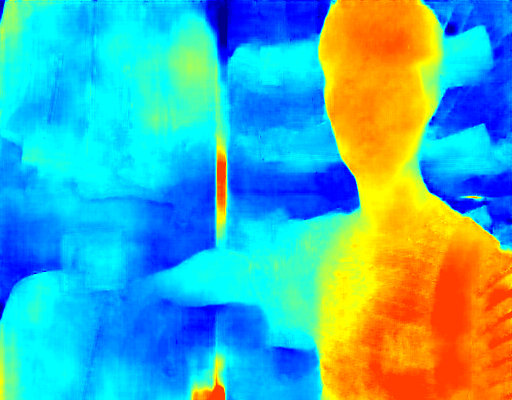}}\hspace{0pt}
        \subfloat[ZOLE]{\includegraphics[width=\figphone]{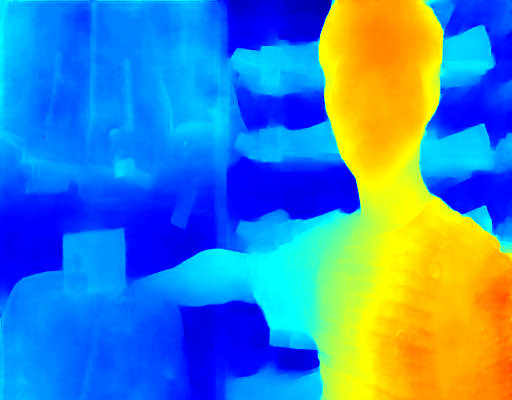}}\\
    \caption{Visual comparisons of several models on the test set of our smartphone data. This figure shows fragments of left images and the corresponding disparity maps obtained with different models. It is clear that our ZOLE approach produces superior disparity results.}
    \label{fig:res_phone}
\end{figure*}
%

\begin{table*}[htbp]
  \centering
  \small
  \caption{Performance comparison of our obtained zoom and learn (ZOLE) model and the other four models.}
    \vspace{2pt}
    \begin{tabular}{c||c:c||c:c|c:c|c:c|c:c|c:c}
    \hline
    \multirow{2}[4]{*}{\bf Dataset} & \multicolumn{2}{c||}{\multirow{2}[4]{*}{\bf Metric}} & \multicolumn{10}{c}{\vspace{-1.7pt}$\phantom{\hat{I}}\mathop{\textbf{Model}}\limits_{\phantom{.}}\phantom{\hat{I}}$} \bigstrut[t]\\
\cline{4-13}          & \multicolumn{2}{c||}{} & \multicolumn{2}{c|}{Tonioni \cite{tonioni2017unsupervised}} & \multicolumn{2}{c|}{{\vspace{-1pt}$\phantom{\hat{I}}\mathop{\textrm{DispNetC}}\limits_{\phantom{.}}\phantom{\hat{I}}$}} & \multicolumn{2}{c|}{DispNetC-80} & \multicolumn{2}{c|}{ZOLE-S} & \multicolumn{2}{c}{ZOLE} \bigstrut[t]\\
    \hline
    \hline
    Smartphone & PSNR  & SSIM  & 22.92 & 0.845 & 21.99 & 0.790 & 22.39 & 0.817 & 22.84 & 0.851 & {\bf 23.12} & {\bf 0.855} \bigstrut[t]\\
    \hline
    FlyingThings3D-80  & EPE   & 3ER   & 1.08  & 6.79\% & 1.03  & 5.63\% & {\bf 0.93}  & {\bf 5.11\%} & 1.10  & 6.88\% & 1.11  & 6.54\% \bigstrut[t]\\
    \hline
    \end{tabular}%
  \label{tab:cmp_mdl}%
  \vspace{-6pt}
\end{table*}%

The Caffe framework \cite{jia2014caffe} is employed to implement our method. 
During training, we randomly crop the images to $640\times 640$ before passing them to a CNN, and let the patch size be $20\times 20$ for building the graphs, resulting in $32 \times 32 = 1024$ graphs for each training example. 
We modify the L1 loss layer of \cite{mayer2016large} to capture the first two terms of \eqref{eq:problem_new}: for a synthetic pair, its L1 loss is weighted by 1.2 times, otherwise the weight is 1. 
We empirically set $w_{\rm left} = 0.3$, $w_{\rm fine} = 0.8$ and $w_{\rm curr} = 1$, all the computed ${\bf s}^{\rm T}_{ij}{\bf L}_{ij}^{(k)}{\bf s}_{ij}$ are averaged then weighted by 1.5 times for a loss (the third term in \eqref{eq:problem_new}). 
We have tried out different up-sampling ratios $r$'s ranging from 1.2 to 2 for computing $D_i$, and found the the obtained CNNs have similar performance. 
In our experiments, we let $r=1.5$. 
Data augmentation is introduced to the synthetic stereo pairs. 
For each individual view in a synthetic pair, Gaussian noise ($\sigma \in\{0, 10, 15\}$) are randomly added. 
The brightness of each image channel are also randomly adjusted (by a factor of $\rho \in\{0.8, 1, 1.2\}$). 
We let the batch size be 6, the learning rate be $5\times 10^{-5}$, and finetune the model for $k_{\rm max}=10^4$ iterations, validation is performed every 500 iterations. 

We first study the following models:
\vspace{-4pt}
\begin{enumerate}[(i)]
    \item ZOLE: Generalize the pre-trained model for smartphone stereo pairs with our method;
    \vspace{-4pt}
    \item ZOLE-S: Remove graph regularization and simply let the CNN iteratively learn its own finer-grain outputs;
    \vspace{-4pt}
    \item DispNetC-80: Finetune the pre-trained model on the FlyingThings3D-80 examples;
    \vspace{-4pt}
    \item DispNetC: Released model pre-trained with FlyingThings3D \cite{mayer2016large}.
\end{enumerate}
\vspace{-4pt}
The very recent method \cite{tonioni2017unsupervised} by Tonioni~{\it et al.} also finetunes a pre-trained model using stereo pairs from the target domain. 
They first estimate disparity maps for the target domain with AD-CENCUS \cite{zabih1994non}. 
To finetune the model, they treat the obtained disparity maps as ``ground-truths'' while taking a confidence measure \cite{poggi2016learning} into account. 
For comparison, we finetune a model with their released code under their recommended settings.

\deflen{figkitti}{118pt}
\begin{figure*}[!t]
        \centering
        \subfloat{\includegraphics[width=\figkitti]{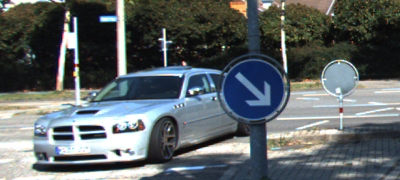}}\hspace{0pt}
        \subfloat{\includegraphics[width=\figkitti]{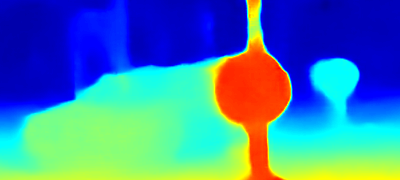}}\hspace{0pt}
        \subfloat{\includegraphics[width=\figkitti]{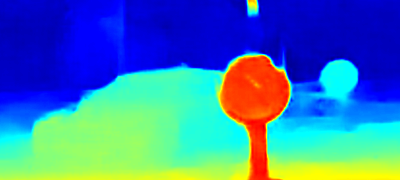}}\hspace{0pt}
        \subfloat{\includegraphics[width=\figkitti]{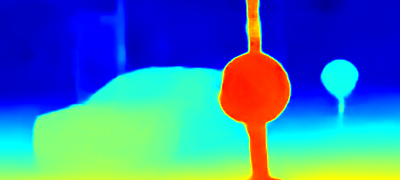}}\\ \vspace{-5pt}
        \subfloat{\includegraphics[width=\figkitti]{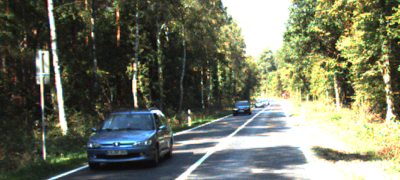}}\hspace{0pt}
        \subfloat{\includegraphics[width=\figkitti]{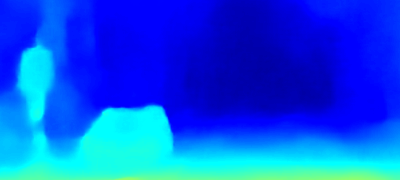}}\hspace{0pt}
        \subfloat{\includegraphics[width=\figkitti]{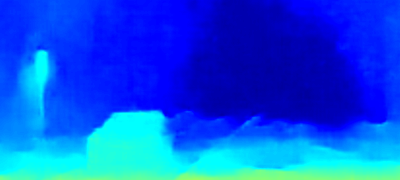}}\hspace{0pt}
        \subfloat{\includegraphics[width=\figkitti]{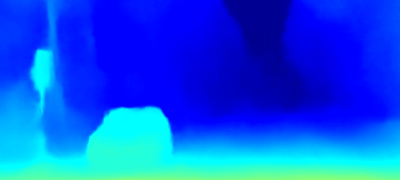}}\\
        \addtocounter{subfigure}{-10}\vspace{-5pt}
        \subfloat[Left image]{\includegraphics[width=\figkitti]{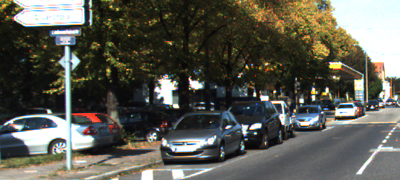}}\hspace{0pt}
        \subfloat[Tonioni {\it et al.}\,\cite{tonioni2017unsupervised}]{\includegraphics[width=\figkitti]{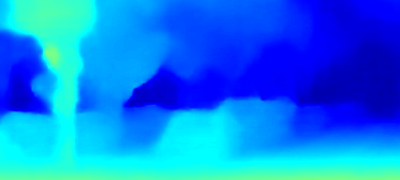}}\hspace{0pt}
        \subfloat[DispNetC]{\includegraphics[width=\figkitti]{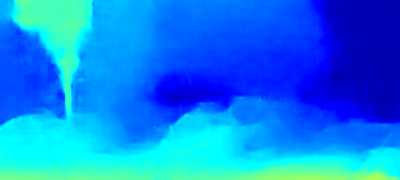}}\hspace{0pt}
        \subfloat[ZOLE]{\includegraphics[width=\figkitti]{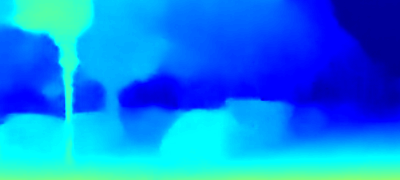}}\\
    \caption{Visual comparisons of several models on the KITTI stereo 2015 dataset, where our ZOLE method produces accurate fine details.}
    \label{fig:res_kitti}
    \vspace{-8pt}
\end{figure*}

Since the stereo pairs of smartphones do not have ground-truth disparities, we evaluate the performance of a model in a way similar to the validation process presented in Section\,\ref{ssec:alg}. 
We synthesize the left images with the estimated disparities and the right images, then measure the difference between the synthesized left images and the genuine ones, using both PSNR and SSIM as the difference metrics. 
For testing or validation, all the stereo pairs are fed to the CNN at a fixed resolution of $1024\times 1024$. 
Figure\,\ref{fig:evo} plots the performance of ZOLE, ZOLE-S and DispNetC-80 on the validation set of the smartphone data during training (measured in terms of average PSNR of the synthesized left images). 
Besides, Table\,\ref{tab:cmp_mdl} presents the performance of all the aforementioned models, on both the test sets of the smartphone data and FlyingThings3D-80. 
We use end-point-error (EPE) and three-pixel error rate (3ER) as the evaluation metrics for the FlyingThings3D-80 dataset. 
Compared to the models trained only with the synthetic data (DispNetC and DispNetC-80), the one obtained with our method (ZOLE) achieves the best PSNR and SSIM performance. 
Figure\,\ref{fig:res_phone} shows visual comparisons of four models on the test sets of the smartphone data. 
One can clearly see that, our approach leads to smooth disparities with very sharp details, while disparity maps produced by other models may be blurry or contain artifacts. 

\begin{table}[t!]
  \centering
  \small
  \caption{Objective performance on the KITTI stereo 2015 dataset.}\vspace{2pt}
    \begin{tabular}{c||c|c|c|c}
    \hline
    \multirow{2}[4]{*}{Metric} & \multicolumn{4}{c}{\vspace{-1.5pt}$\phantom{\hat{I}}\mathop{\textrm{Model} }\limits_{\phantom{.}}\phantom{\hat{I}}$} \bigstrut[t]\\
\cline{2-5}          & Tonioni\,\cite{tonioni2017unsupervised} & DispNetC & ZOLE-S & ZOLE \bigstrut[t]\\
    \hline
    EPE   & 1.27  & 1.64  & 1.34  & {\bf 1.25} \bigstrut[t]\\
    3ER   & 7.06\% & 11.41\% & 7.56\% & {\bf 6.76}\% \bigstrut[t]\\
    \hline
    \end{tabular}%
  \label{tab:cmp_kti}%
  \vspace{-10pt}
\end{table}%
\subsection{Driving Scenes of KITTI}
Our self-adaptation method is also applied to generalize the pre-trained DispNetC model to the KITTI stereo 2015 dataset \cite{geiger2012we}, which contains dynamic street views from the perspective of a driving car. 
The KITTI stereo 2015 dataset have 800 stereo pairs. 
Among them, 200 examples have publicly available (sparse) ground-truth disparity maps. 
They are employed for testing, while the rest 600 pairs are used for validation. 
For training, we first gather $N_{\rm syn}=9000$ stereo pairs randomly from the FlyingThings3D dataset. 
Since the KITTI 3D object 2017 dataset \cite{geiger2012we} have more than 10k stereo pairs of the same characteristics as KITTI stereo 2015, we randomly pick $N_{\rm dom}=3000$ stereo pairs from it for training. 
During training, we adopt similar settings as presented in Section\,\ref{ssec:exp_phone}. 
However, in this scenario, all images are resized to $1280\times 400$ then randomly cropped to $1024\times 384$ before passing to the CNN for training. 

We hereby compare our approach, ZOLE, with models obtained with ZOLE-S and \cite{tonioni2017unsupervised}; while the original DispNetC model is adopted as a baseline. 
For a fair comparison, all the images are resized to $1280\times 384$ before feeding to the network. 
Table\,\ref{tab:cmp_kti} presents the objective metrics of ZOLE, along with those of the competing methods.
We see that our method has the best objective performance, while the method of Tonioni~{\it et~al.} also provides a reasonable gain. 
Figure\,\ref{fig:res_kitti} shows several fragments of the resulting disparity images. 
One can see that our method provides accurate edges even for very fine details.

More results and discussions are provided in the supplementary material. 
Our method is essentially different from those deep stereo algorithms relying on left-right consistency for backpropagation \cite{zhong2017self,zhou2017unsupervised}. 
Hence, it is possible to combine our rationale---discriminatively learns from the finer-grain outputs---with these methods to achieve further improvements. 
Moreover, the same rationale can be applied to other pixel-wise regression/classification problems, {\it e.g.}, optical flow estimation \cite{ilg2017flownet,mayer2016large} and segmentation \cite{long2015fully}. 
We leave these research directions for future exploration.

\section{Conclusion}
\label{sec:conclusion}
Due to the deficiency of ground-truth data, it is difficult to generalize a pre-trained deep stereo model to a novel domain. 
To tackle this problem, we propose a self-adaption approach for CNN training without ground-truth disparity maps of the target domain. 
We first observe and analyze two phenomena, namely, generalization glitches and scale diversity. 
To exploit scale diversity while avoiding generalization glitches, we let the model learn from its own finer-grain output, while a graph Laplacian regularization is imposed to selectively keep the desired edges and smoothing out the artifacts. 
We call our method zoom and learn, or ZOLE for short. 
It is applied to two domains: daily scenes collected by smartphone cameras and street views captured from the perspective of a driving car.

{\small
\bibliographystyle{ieee}

}

\end{document}